\documentclass[letterpaper, 10 pt, conference]{ieeeconf}  

\IEEEoverridecommandlockouts                              

\usepackage{graphics} 
\usepackage{epsfig} 
\usepackage{mathptmx} 
\usepackage{times} 
\usepackage{amsmath} 
\usepackage{amssymb}  
\usepackage{subfig}
\usepackage{tabularx}
\usepackage{booktabs}
\usepackage{hyperref}
\usepackage{fancyhdr}

\title{\LARGE \bf
Swim: A General-Purpose, High-Performing, and Efficient Activation Function for Locomotion Control Tasks}

\author{Maryam Abdool$^{1}$ and Tony Dear$^{2}$
\thanks{$^{1}$Maryam Abdool is with Department of Computer Science,
        Columbia University, New York, NY 10027, USA
        {\tt\small maryam.r.abdool@gmail.com}}%
\thanks{$^{2}$Tony Dear is with Department of Computer Science,
        Columbia University, New York, NY 10027, USA
        {\tt\small tony.dear@columbia.edu}}%
}

\fancypagestyle{FirstPage}{
\lfoot{Manuscript submitted for review to the IEEE/RSJ International Conference on Intelligent Robots and Systems (IROS)} 
}

\begin{document}

\onecolumn 
\begin{center}

\vspace*{\stretch{1}}

  \huge\bfseries 
  Notice \\
  \LARGE\bfseries 

  This work has been submitted to the IEEE for possible publication. Copyright may be transferred without notice, after which this version may no longer be accessible.
\vspace*{\stretch{1}}
\end{center}
\twocolumn 

\maketitle
\thispagestyle{empty}
\pagestyle{empty}

\thispagestyle{FirstPage}

\begin{abstract}

    Activation functions play a significant role in the performance of deep learning algorithms. In particular, the Swish activation function tends to outperform ReLU on deeper models, including deep reinforcement learning models, across challenging tasks. Despite this progress, ReLU is the preferred function partly because it is more efficient than Swish. Furthermore, in contrast to the fields of computer vision and natural language processing, the deep reinforcement learning and robotics domains have seen less inclination to adopt new activation functions, such as Swish, and instead continue to use more traditional functions, like ReLU. To tackle those issues, we propose Swim, a general-purpose, efficient, and high-performing alternative to Swish, and then provide an analysis of its properties as well as an explanation for its high-performance relative to Swish, in terms of both reward-achievement and efficiency. We focus on testing Swim on MuJoCo’s locomotion continuous control tasks since they exhibit more complex dynamics and would therefore benefit most from a high-performing and efficient activation function. We also use the TD3 algorithm in conjunction with Swim and explain this choice in the context of the robot locomotion domain. We then conclude that Swim is a state-of-the-art activation function for continuous control locomotion tasks and recommend using it with TD3 as a working framework.

\end{abstract}

\section{INTRODUCTION AND RELATED WORK}

    Deep reinforcement learning can automate the design of complex controllers for locomotion tasks \cite{haarnoja2018learning}. Before the emergence of deep reinforcement learning, a common technique was to manually design controllers for each locomotion task; this process requires an accurate dynamic model of the robot that may be difficult to achieve \cite{haarnoja2018learning}. In addition to exhibiting complex dynamics, locomotion problems also feature high degrees of freedom, complicating the task of engineering controllers \cite{duan2016benchmarking}. Therefore, we utilize deep reinforcement learning to automate the design of our controllers for MuJoCo’s locomotion continuous control tasks presented in this paper.
    
    In a neural network, each neuron performs a transformation on the inputs using the incoming weights and biases, but such a simple stand-alone model fails to learn complex tasks \cite{schmidhuber2015deep}. Therefore, non-linearity is introduced so that neural networks can act as non-linear function approximators in various settings, including reinforcement learning \cite{sutton2018reinforcement}. This is achieved by using non-linear activation functions, such as the Rectified Linear Unit (ReLU), in the hidden layers of the neural network \cite{nair2010rectified}. Non-linearity is a powerful characteristic of neural networks because of the universal approximation theorem’s implication that non-linear activation functions can approximate any continuous function arbitrarily well \cite{hornik1989multilayer}. This has vital implications for the application of deep reinforcement learning algorithms to continuous control and locomotion tasks, as it guarantees that neural networks have the potential to learn complex controllers.
    
    TD3 is one such deep reinforcement learning algorithm that is designed to learn complex locomotion controllers \cite{fujimoto2018addressing}. As a deep learning algorithm, TD3 follows the standard approach of using ReLU, a non-linear activation function, in the hidden layers of the neural network to model complex tasks \cite{fujimoto2018addressing}. What distinguishes TD3 from other algorithms are three aspects: an actor-critic approach, sample efficiency, and the twin critic networks. For these reasons, we select TD3 over other algorithms.
    
    First, in TD3’s actor-critic neural network architecture, the policy network is called the actor because it is used to select actions, and the estimated value function network is called the critic because it criticizes the actions made by the actor \cite{sutton2018reinforcement}. An actor-critic method is advantageous because it can be applied to problems with continuous action spaces, namely MuJoCo’s continuous control tasks, where Q-learning methods cannot be directly applied; for continuous control tasks with an infinite set of actions, Q-learning methods must search through the infinite set to select the action, while the separation of the policy from the value function in the actor-critic method reduces the extensive computation needed for each action selection \cite{sutton2018reinforcement}. Second, TD3 is sample-efficient because it is based on a deterministic policy gradient method \cite{silver2014deterministic}. The deterministic policy gradient has the simple form of the expected gradient of the action-value function \cite{silver2014deterministic}. Because of this simple form, the deterministic policy gradient can be estimated much more efficiently than its stochastic counterpart \cite{silver2014deterministic}. Finally, TD3 uses two critic networks, instead of one, to address the overestimation bias problem present in DDPG and other Q-learning algorithms that occurs when the noisy value estimate is maximized \cite{fujimoto2018addressing}, \cite{thrun1993issues}.
    
    Although designing new algorithms is an important problem, such as the TD3 algorithm that we employ to model our controllers, research in creating new non-linear activation functions is neglected in favor of designing more complex deep reinforcement learning algorithms. Therefore, we pursue the field of activation functions as research shows that not all non-linear activation functions have the same performance. For example, it is believed that ReLU is advantageous over Sigmoid because it reduces the likelihood of the vanishing gradient problem and it is a simpler, more efficient activation function \cite{glorot2011deep}. Similarly, \cite{ramachandran2017searching} speculates that Swish tends to outperform common baseline functions, including ReLU, because it is non-monotonic. However, Swish is less efficient than ReLU due to the required exponential computations, which is one of the reasons ReLU is a widely used function, in addition to being more well-known; as mentioned earlier, the choice of using ReLU in the original TD3 implementation is one notable example \cite{howard2019searching}, \cite{fujimoto2018addressing}. 
    
    Because of those limitations, we invent Swim, a non-monotonic, smooth activation function mathematically defined as $f(x) = {\frac{x}{2}}({\frac{kx}{\sqrt{1+k^2x^2}}+1})$. Swim is more efficient and more likely to be high-performing than Swish, and is consequently also more likely to be high-performing than ReLU; as mentioned earlier, \cite{ramachandran2017searching} and \cite{elfwing2018sigmoid} have shown that Swish tends to outperform ReLU on deeper models, including deep reinforcement learning models, across challenging tasks. We achieve efficiency and high performance without compromising any of the desirable properties, such as smoothness, that make Swish outperform baseline functions. This is in contrast to previous work, such as \cite{howard2019searching}'s design of H-Swish, an efficient but non-smooth version of Swish. Ultimately, we strive to achieve high performance and efficiency as better activation functions correspond to better locomotion controllers. High performance and efficiency are important metrics for the locomotion domain due to the complex dynamics and time-intensive characteristics mentioned earlier. Finally, we also hypothesize that there are specific properties intrinsic to Swim’s success in the deep reinforcement learning and locomotion control domain.

\section{SWIM: AN ANALYSIS}

	The newly proposed activation function, Swim, and its first derivative are defined as:

	\begin{equation} 
		\label{one}
            \begin{gathered}
		f(x) = {\frac{x}{2}}\left({\frac{kx}{\sqrt{1+k^2x^2}}+1}\right), \\ 
      f'(x) = \frac{1}{2}\left({\frac{kx(k^2x^2 + 2)}{(\sqrt{1+k^2x^2})^3}+1}\right)
        \end{gathered}
	\end{equation}

    where ${k}$ is a constant that can be tuned or defined before training. We pick a ${k}$ such that it can support our analysis that Swim outperforms Swish because of the properties intrinsic to it, and not because of a constant rescaling of the function that could be analogously applied to Swish by changing $\beta$, which is also a learnable parameter. For this reason, we set $k$ = 0.5 to approximate the Swish function ($x\frac{1}{1+e^{-{\beta}x}}$) at $\beta$ = 1, which is the parameter that the original Swish paper runs and reports the results on \cite{ramachandran2017searching}. This also means that this may not be the most optimal setting of the $k$ constant, but we make this trade-off to support our analysis. The difference between the two functions, alongside ReLU ($max(0,x)$) \cite{nair2010rectified}, can be seen in Figure 1.

	\begin{figure}[h!]
  		\subfloat[]{\includegraphics[width=\linewidth, height=44mm]{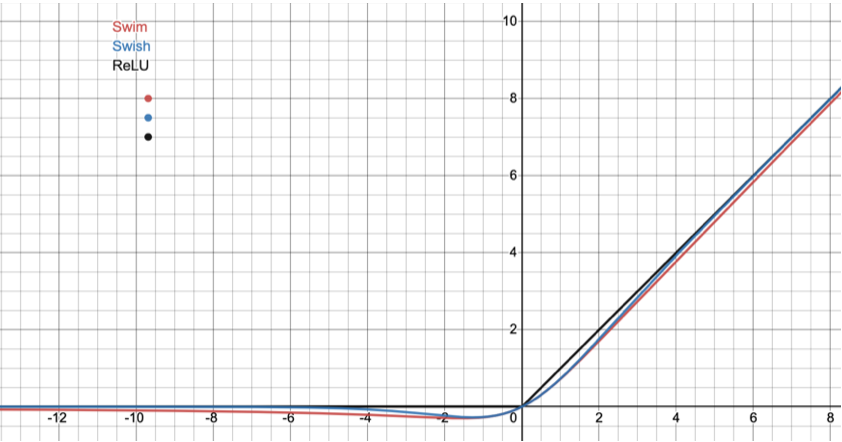}}
        \hspace{.01\linewidth}
  		\subfloat[]{\includegraphics[width=\linewidth, height=44mm]{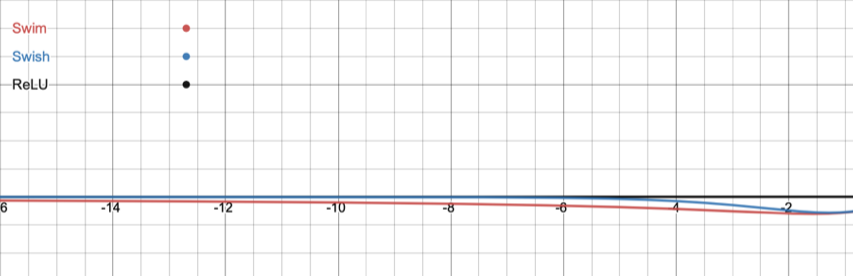}}
  		\caption{(a) Graphs of Swim, Swish, and ReLU. (b) Zoomed-in graphs of Swim, Swish, and ReLU at the negative values.}
  		\label{fig1}
	\end{figure}
 
	Similarly, the plots of the first derivatives of Swim, Swish, and ReLU are shown in Figure 2.

	\begin{figure}[h!]
  		\subfloat[]{\includegraphics[width=\linewidth, height=44mm]{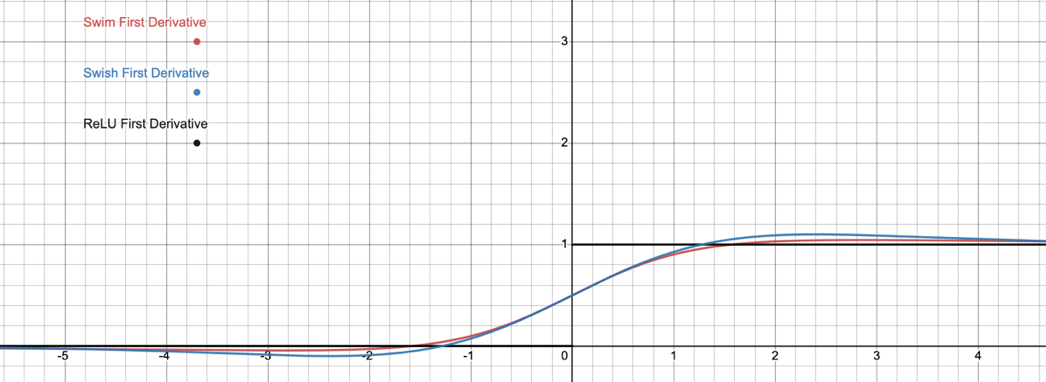}}
        \hspace{.01\linewidth}
  		\subfloat[]{\includegraphics[width=\linewidth, height=44mm]{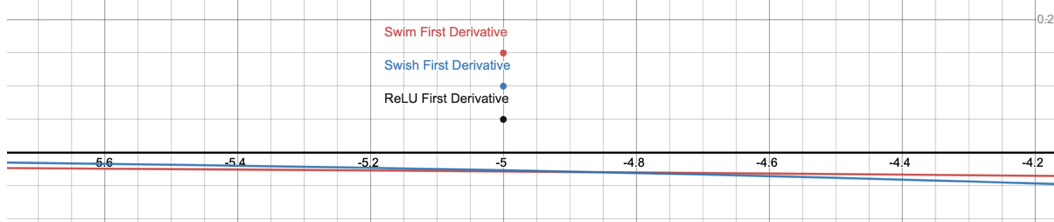}}
  		\caption{(a) First derivative graphs of Swim, Swish, and ReLU. (b) Zoomed-in first derivative graphs of Swim, Swish, and ReLU at the negative values.}
  		\label{fig2}
	\end{figure}

	In the next subsection, we incorporate the behaviors of Swim, Swish, and ReLU and their first derivatives in our analysis. We focus our analysis on comparing Swim (for ${k}$ = 0.5) against Swish as \cite{ramachandran2017searching} and \cite{elfwing2018sigmoid} have shown that Swish/SiLU outperforms ReLU and other baseline functions across various tasks.

    \subsection{Properties of Swim}
    Like Swish, Swim is unbounded above, bounded below, non-monotonic, and smooth. Out of these four properties, non-monotonicity and smoothness are the properties that distinguish Swish and Swim from ReLU and other baseline activations. As shown in Figure 1, the non-monotonicity property allows Swim and Swish to produce negative values for small negative inputs. The impact on the gradient is shown in Figure 2, where the first derivative of Swim and Swish are larger than zero for those small negative inputs. This has the effect of increasing expressivity and improving gradient flow \cite{ramachandran2017searching}. Similarly, the smoothness of Swim and Swish correspond to a smooth loss landscape that is easier to optimize \cite{ramachandran2017searching}.
    
    As shown in Figure 1, negative inputs of Swim saturate at a slower rate than Swish. For ${k}$ = 0.5 (Swim) and $\beta$ = 1 (Swish), Swim maintains this property while bounding the Swish function from below for all negative inputs, noting that both functions have an approximately equal local minimum value of ~0.3 in the negative region. Essentially, this allows Swim to produce more and larger negative values for negative inputs without causing the gradient to explode, and without forgoing the regularization effect that is induced for functions that approach zero \cite{glorot2011deep}. Although the value of Swish’s $\beta$ parameter could be decreased to match Swim’s slow saturation rate, it cannot be decoupled from the consequence of making Swish approach a linear function, which depending on some tasks, may be undesirable as well as disadvantageous due to losing the properties of a non-linear activation function. 
    
    In the context of deep off-policy temporal difference algorithms, namely TD3, Swim’s smoothly-saturating property becomes relevant as the neural networks do not incorporate feature normalization techniques. This results in a non-zero-centered (unnormalized) activation function that becomes susceptible to experiencing the vanishing gradient problem at points that saturate quickly, which is the case for Swish. In other words, Swim mitigates the lack of normalization issue in TD algorithms by slowly saturating the negative inputs. As \cite{bhatt2019crossnorm} has suggested, normalization techniques are not used partly because the action-value function is evaluated two times  $({Q(s,a)}$ and $Q(s',\pi(s')))$ during the training of the critic network, thereby producing actions that come from different distributions. Furthermore, the observation space is also not necessarily normalized, which is the case for the MuJoCo locomotion continuous control benchmark tasks on which we run our tests. 
    
    Unlike Swish, Swim is less likely to overflow due to the absence of the exponentials \cite{blanchard2021accurately}. As shown in Equation 1, Swim and its first derivative are instead composed of square roots and a quadratic term. Furthermore, Swim is also less computationally expensive than Swish. Although they are asymptotically equivalent functions, in practice, the square root and quadratic functions are faster to compute than exponentials, and the inverse square roots as a whole, which the Swim function and its derivative are composed of, are even faster to calculate than exponentials \cite{carlile2017improving}. John Carmack is regarded to have written a fast implementation of the inverse square root function \cite{lomont2003fast}, and on an x86 architecture, \cite{carlile2017improving} has shown that his inverse square root-based activation function is faster than \cite{clevert2015fast}'s exponential-based activation function named ELU. Furthermore, Intel publishes the CPU performance of vector functions, including the Inverse Square Root and the Exponential/Exp function, which \cite{carlile2017improving} has aggregated as shown in Table I. Depending on the x86 architecture, Table I shows that the Inverse Square Root is approximately 1.2-2.2x faster than the Exp function.
    
    	\begin{table}[h!]
		\caption{CPU Performance of the Vector Inverse Square Root and Exponential Functions (Clocks per Element)}
		\label{tabletest}
		\begin{center}
  			\begin{tabularx}{\linewidth}{ X X X X }
    				\toprule
    				Vector Function Single Precision (EP) & Intel Xeon E5-2699 v3 (Haswell AVX2) & Intel Xeon E5-2699 v4 (Broadwell AVX2) & 
    				Intel Xeon Platinum 8180 (Skylake AVX-512)\\ 
    				\midrule
    				InvSqrt & 0.66   & 0.64 & 0.24 \\ 
    				Exp & 0.81 & 0.89 & 0.52 \\ 
    				Exp/InvSqrt & 1.2× & 1.4× & 2.2× \\
    				\bottomrule
  			\end{tabularx}
		\end{center}
	\end{table}

	\section{METHODS}
	
		\begin{figure}[h!]
  		\subfloat[]{\includegraphics[width=.5\linewidth, height=44mm]{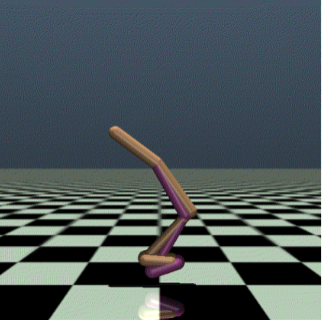}}
  		\subfloat[]{\includegraphics[width=.5\linewidth, height=44mm]{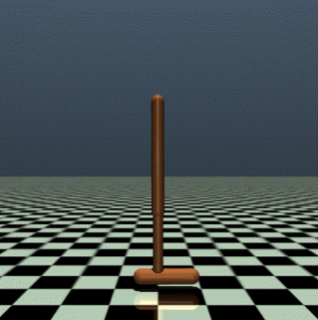}} \\ 
  		\subfloat[]{\includegraphics[width=.5\linewidth, height=44mm]{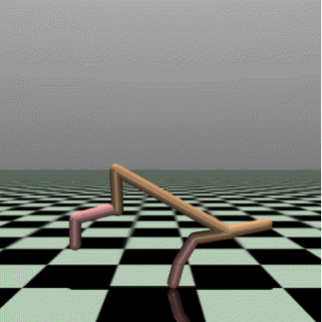}}
  		\subfloat[]{\includegraphics[width=.5\linewidth, height=44mm]{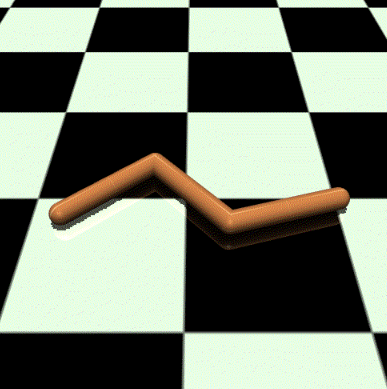}}
  		\caption{(a) Walker2d-v2. (b) Hopper-v2. (c) HalfCheetah-v2. (d) Swimmer-v2.}
  		\label{fig3}
	\end{figure}

    Shown in Figure 3, we run our experiments on four of MuJoCo’s benchmark locomotion control environments: (a) Walker2d-v2, (b) Ant-v2, (c) Hopper-v2, and (d) HalfCheetah-v2. We use the TD3 algorithm and the Swim activation function to solve those environments. We also report the performance and efficiency of those experiments.

    \subsection{Preliminaries}
    We model each environment as a Markov Decision Process (MDP), where at each discrete time step ${t}$ and a given state $s \in S$, an agent selects an action $a \in A$ according to its policy $\pi : S \to A$. The agent then receives a reward $r$ and the new state of the environment $s'$. The objective is to then find the optimal policy $\pi_\phi$, with parameters $\phi$. 

    We select the Twin Delayed Deep Deterministic Policy Gradient algorithm (TD3) \cite{fujimoto2018addressing} to learn the optimal policy. Based on the Deep Deterministic Policy Gradient (DDPG) algorithm \cite{lillicrap2015continuous}, TD3 maintains a single actor and two critic networks, with the action-value functions, $Q_{\phi_{1}}(s,a)$ and $Q_{\phi_{2}}(s,a)$, acting as the two critic neural network approximators. TD3 learns the two Q-functions, $Q_{\phi_1}$ and $Q_{\phi_2}$, by mean square Bellman error minimization.
    
    At each time step, the targets in the Bellman error loss functions are updated according to the following equation:

	\begin{equation} 
		\label{two}
            \begin{gathered}
            y = r + \gamma \min_{i=1,2} Q_{\theta_{i}'} (s', \pi_{\phi'} (s') + \epsilon), \\
            \epsilon  \sim clip(N (0, \sigma ), -c, c)
            \end{gathered}
	\end{equation}
    
	where $\gamma$ is the discount factor, ${\theta_{i}'}$ is the parameter of the target Q networks, ${\phi'}$ is the parameter of the target actor network, $\epsilon$ is the clipped noise, and $(-c, c)$ are hyperparameter values that are set to clip target policy noise. That is, the smaller of the two target values given by the Q-functions is used to update the target. 
	
    We use the authors’ original implementation of TD3 in PyTorch because the original paper also focuses on running TD3 on MuJoCo control tasks, including the ones we use in this paper. Therefore, we also use the same optimal hyperparameters and only change the activation functions in our experiments. 
    
    \subsection{Applying Activation Functions}
    We write our custom implementation of the Swim activation function in PyTorch and use it in all of the hidden layers of the actor and critic networks. We follow the same approach when applying the Swish activation function, although Swish uses the underlying numerically stable Sigmoid torch function in PyTorch. We do not implement a numerically stable version but write a direct translation of the Swim function, as written in Equation 1, using the torch functions. A numerically stable function could potentially improve the performance of Swim, although we leave that for future work. We also use double-precision instead of single-precision to improve accuracy and run our experiments on a machine with an ARM architecture. 

    As mentioned in Section II, we only evaluate Swim against Swish as \cite{ramachandran2017searching} and \cite{elfwing2018sigmoid} have shown that Swish/SiLU outperforms ReLU and other baseline functions across various tasks. We also select Swim as an activation function for this problem for reasons explained in Section I and Section II.A, and also because simple kinematic controller models, such as $tanh(x)$ \cite{xie2018trajectory}, would not suffice as an activation function for this problem since locomotion control tasks exhibit complex dynamics. 
    
    \subsection{Evaluation}
    We use performance curves to compare the performance of Swim with Swish, following standard practices in deep reinforcement learning. We also measure inference time of the actor network to compare the efficiency of the two activation functions, effectively isolating the evaluation of efficiency from performance.

    \section{RESULTS}
    For the performance metric, we run TD3 with the selected activation function for 1 million time steps on each environment, where we record the average reward over 10 episodes with no exploration noise every 5000 time steps. We then report the performance curves over 5 random seeds for each environment. We also report the max average return over those 5 seeds to compare the results of the activation functions.
    
    For the efficiency metric, we calculate the inference time over the first 100 iterations on each environment and report the results. 
    
    \subsection{Performance}

    \begin{figure}[h!]
  		\subfloat[]{\includegraphics[width=.5\linewidth, height=44mm]{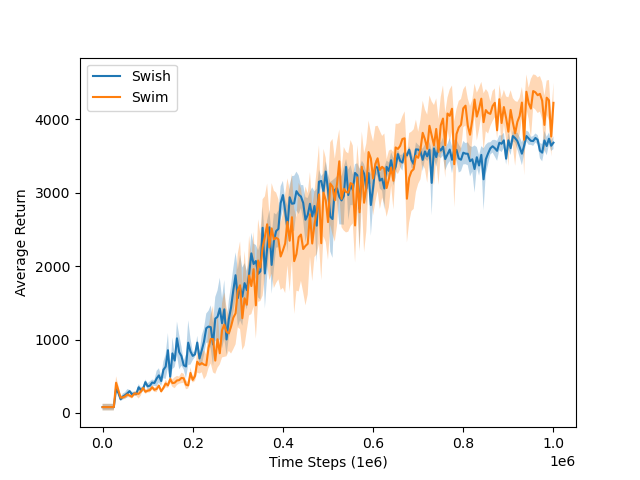}}
  		\subfloat[]{\includegraphics[width=.5\linewidth, height=44mm]{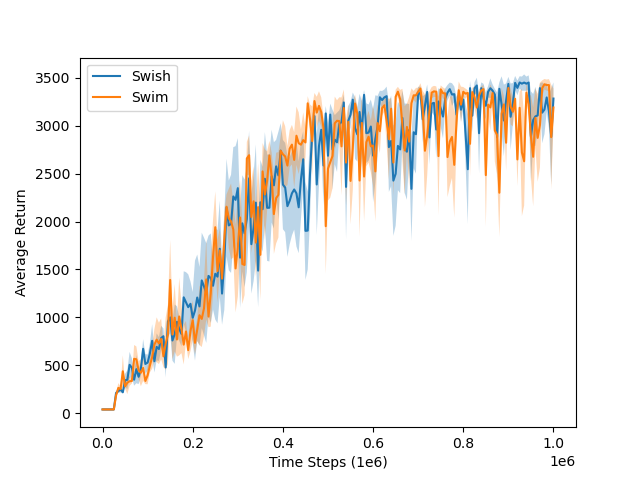}} \\
  		\subfloat[]{\includegraphics[width=.5\linewidth, height=44mm]{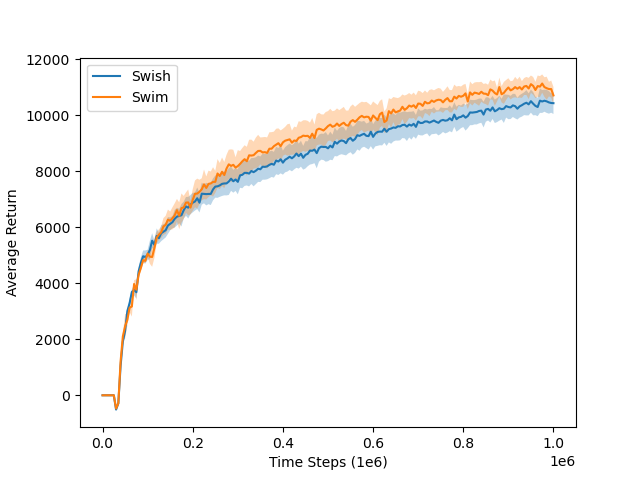}}
  		\subfloat[]{\includegraphics[width=.5\linewidth, height=44mm]{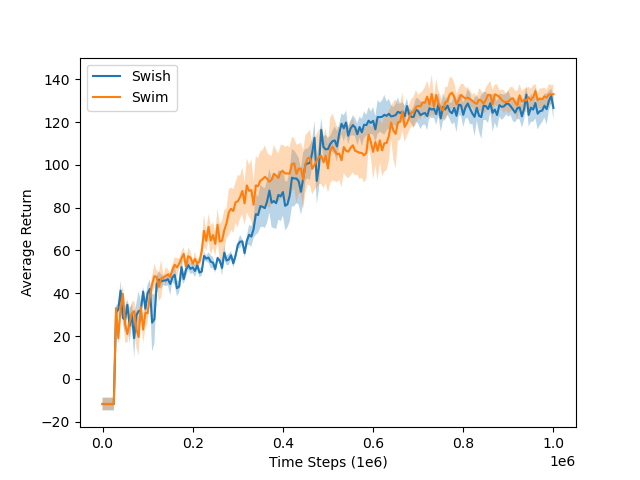}}
  		\caption{ Learning curves for MuJoCo's continuous control tasks: (a) Walker2d-v2, (b) Hopper-v2, (c) HalfCheetah-v2, and (d) Swimmer-v2. The shaded region represents half a standard deviation of the average evaluation over 5 trials.}
  		\label{fig4}
	\end{figure}

    \begin{table}[h!]
		\caption{Max Average Return over 5 trials of 1 million time steps. ± corresponds to a single standard deviation over trials.}
		\label{tabletest2}
		\begin{center}
  			\begin{tabularx}{\linewidth}{ X X X X }
    				\toprule
    				Environment & Swish & Swim & Improvement\\ 
    				\midrule
    				Walker2d-v2 & 3774.668 ± 149.779 & 4387.442 ± 453.526 & 16\% \\
    				Hopper-v2 & 3450.055 ± 129.203 & 3431.081 ± 110.394	& 0\% \\ 
    				HalfCheetah-v2 & 10519.091 ± 872.693 & 11130.364 ± 652.716 & 5\% \\
                    Swimmer-v2 & 132.838 ± 8.146 & 134.503 ± 6.806 & 1\% \\
    				\bottomrule
  			\end{tabularx}
		\end{center}
	\end{table}
    
    The learning curves are presented in Figure 4 with the max average return reported in Table II. The results show that Swim matches or outperforms Swish in both final performance and learning speed across all tasks. Based on the curve and the max average return of the two activation functions, Swim outperforms Swish in 3 out of 4 environments: the Walker2d-v2, HalfCheetah-v2, and Swimmer-v2; however, improvement is more significant in the Walker2d-v2 and HalfCheetah-v2 environments, with a 16\% and 5\% improvement rate, respectively. 

    The results support the hypothesis and analysis presented in Section II; for $k$ = 0.5, it is expected that Swim will not always outperform Swish, as the constant has been selected such that Swim will approximate Swish at $\beta$ = 1. An exploration of a different value for $k$ could potentially result in higher performance.
    
    However, Swim still outperforms Swish in some environments. As explained in Section II, we theorize that this is because the negative values of Swim saturate at a slower rate than Swish, which mitigates the issue of the lack of feature normalization techniques in TD3 and other deep reinforcement learning algorithms, in addition to the fact that the observation space of our environments is not normalized. 

    \subsection{Efficiency}

    \begin{table}[h!]
		\caption{Inference time over the first 100 iterations for each environment. ± corresponds to a single standard deviation over the iterations.}
		\label{tabletest3}
		\begin{center}
  			\begin{tabularx}{\linewidth}{ X X X X }
    				\toprule
    				Environment & Swish (ms) & Swim (ms) & Improvement\\ 
    				\midrule
    				Walker2d-v2 & 0.810 ± 0.066	& 0.687± 0.077 & 17.9\% \\
    				Hopper-v2 & 0.720 ±  0.055	& 0.666 ± 0.056	& 8.1\% \\ 
    				HalfCheetah-v2 & 0.826 ± 0.057	& 0.733 ± 0.064	& 12.7\% \\
                    Swimmer-v2 & 0.737 ± 0.060	& 0.644± 0.045 & 14.4\% \\
    				\bottomrule
  			\end{tabularx}
		\end{center}
	\end{table}
 
    Inference time for each of the four environments is reported in Table III. The table shows that Swim consistently outperforms Swish in terms of efficiency across all environments, with improvement ranging from 8.1\% to 17.9\%. In contrast to the results of the max average return, improvement in efficiency is significant across all environments.

    The results support our analysis in Section II, where we hypothesize that Swim is more efficient than Swish because inverse square roots are computationally less expensive than exponentials. Since we did not implement a custom backward pass that leverages the properties of Swim, we do not measure backpropagation time, although the inverse square root is defined in the first derivative of the Swim function, as shown in Equation 1, which could be leveraged to optimize the efficiency of the backward pass as \cite{carlile2017improving} has done in his paper.

    \section{DISCUSSION AND FUTURE WORK}
    Based on our analysis and results, the primary advantage of Swim is its reduced computational complexity compared with Swish, as it significantly outperforms Swish in efficiency across all tasks, although it also achieves state-of-the-art performance and sometimes exceeds Swish in learning speed and reward-achievement. Swim’s low computational complexity allows it to be a suitable general-purpose activation function, which is why we urge future researchers to test and optimize Swim on other models beyond deep reinforcement learning and the robotics domain. For the deep reinforcement learning and robotics domain, Swim is especially applicable because of its both high-performing and high-efficiency properties, where locomotion continuous control tasks benefit from deep off-policy temporal difference learning algorithms that use activation functions with negatively saturating inputs to mitigate the lack of feature normalization techniques in addition to the lack of normalization of the observation space; locomotion continuous control tasks also benefit from an efficient activation function that accelerates running time because they exhibit complex dynamics and high degrees of freedom, requiring them to take a long time in training. Those results also give rise to the idea of domain-specific activation functions, which current research is not focused on; current research trends instead focus on designing generalizable activation functions and then testing them on various tasks, without investigating the properties that would benefit certain areas more than others. We hope to see future work in the area of domain-specific activation functions, especially in the fields of deep reinforcement learning and robotics that see less work with activation functions. 
        
    Further work with Swim remains. In terms of testing and optimization, we only test our function on the CPU, but not on the GPU. This also means that Swim could be optimized using a CUDA-based implementation to further improve efficiency as \cite{misra2019mish} has done in his paper. We also only test our results on an ARM architecture and report our efficiency results, although \cite{carlile2017improving} has previously reported the fast results of the inverse square root on an x86 architecture, as well as Intel has published the fast performance of the inverse square root, which we have reported in this paper. Furthermore, the backward pass of Swim could be implemented such that its efficiency is improved using the inverse square root as \cite{carlile2017improving} has done in his paper. In terms of the activation function itself, we do not implement a numerically stable version, although the performance we report shows that Swish matches or outperforms Swish in learning speed and reward-achievement without doing so. We speculate that a numerically stable version would further improve the performance of Swim. We also speculate that a different value of $k$ could potentially improve the performance of Swim. Based on these findings, we plan on exploring and improving the performance and efficiency of Swim by testing it on a GPU and an ARM architecture, writing a CUDA-based implementation, designing a custom backward pass for Swim, inventing a numerically stable version, and exploring different values of $k$.

    \section*{APPENDIX}

    The code and other supplementary materials used for this research paper, including our own implementation of Swim, are available at \href{https://github.com/maryam-abdool/Swim-Control}{https://github.com/maryam-abdool/Swim-Control}

\addtolength{\textheight}{-12cm}   





\bibliographystyle{plain} 
\bibliography{root}

\end{document}